\documentclass[conference]{IEEEtran}

\usepackage{graphicx}
\usepackage{amsmath}
\usepackage{cite}
\usepackage{url}
\usepackage{booktabs}

\begin{document}

\title{Real Time NILM Based Power Monitoring of Identical Induction Motors Representing Cutting Machines in Textile Industry}

\author{
\IEEEauthorblockN{
Md Istiauk Hossain Rifat*, Moin Khan, Zohara Kamal, Md Borhan Uddin Khan, Mohammad Zunaed
}
\IEEEauthorblockA{
Department of Electrical and Electronic Engineering, BRAC University, Dhaka, Bangladesh\\
}
}

\maketitle

\begin{abstract}
The textile industry in Bangladesh is one of the most energy-intensive sectors, yet its monitoring practices remain largely outdated, resulting in inefficient power usage and high operational costs. To address this, we propose a real-time Non-Intrusive Load Monitoring (NILM)-based framework tailored for industrial applications, with a focus on identical motor-driven loads representing textile cutting machines. A hardware setup comprising voltage and current sensors, Arduino Mega and ESP8266 was developed to capture aggregate and individual load data, which was stored and processed on cloud platforms. A new dataset was created from three identical induction motors and auxiliary loads, totaling over 180,000 samples, to evaluate the state-of-the-art MATNILM model under challenging industrial conditions. Results indicate that while aggregate energy estimation was reasonably accurate, per-appliance disaggregation faced difficulties, particularly when multiple identical machines operated simultaneously. Despite these challenges, the integrated system demonstrated practical real-time monitoring with remote accessibility through the Blynk application. This work highlights both the potential and limitations of NILM in industrial contexts, offering insights into future improvements such as higher-frequency data collection, larger-scale datasets and advanced deep learning approaches for handling identical loads.
\end{abstract}


\section{Introduction}
Bangladesh’s textile industry is one of the most significant contributors to its economy, accounting for over 80\% of export earnings and approximately 10\% of the national GDP \cite{diversegy2025}. As a sector highly dependent on electrical energy for its production processes such as spinning, weaving, dyeing and finishing it ranks among the top ten most energy-intensive industries in the country \cite{bgmea2025}. The garments and textile sectors, which contribute over 80\% of Bangladesh's export earnings and over 10\% of the country's GDP, can enhance aggregate energy efficiency by 25-31\% \cite{lawrence2019}. Moreover, several studies reveal that the sector as a whole holds significant untapped energy efficiency potential, ranging from 25\% to 31\%, which highlights the urgency of adopting monitoring and optimization solutions \cite{tbs2023}. Energy efficiency is vital for Bangladesh's industries, especially when meeting their global clients' ever-increasing environmental, social and governance  requirements.

Despite its economic importance, energy consumption in most textile plants, particularly among small and medium-sized enterprises , is managed through outdated or manual practices. This results in increased production costs, inefficient energy use and missed opportunities for sustainable development. Compounding this issue, many factories rely on captive power or backup generators to maintain production during electricity instability, leading to higher costs and further inefficiencies \cite{lightcastle2024}. Manual monitoring systems are not effective in providing real-time data and human error often affects the accuracy of energy usage reports\cite{europeanparliament2024}. This lack of reliable monitoring leads to undetected power waste, faulty machine operation and high operational costs\cite{fibre2fashion2025}. To meet both environmental targets and buyer demands for sustainable production, textile industries need a more intelligent, low-cost and automated monitoring solution\cite{tbs2025a}.

By enabling real-time visibility and analysis of energy use across individual machines and processes, this system aims to help factories reduce wastage, forecast load demand and align with national and global goals for energy efficiency and greenhouse gas (GHG) reduction\cite{lu2025}. Furthermore, real-time IoT-based monitoring has already been shown in other industries to optimize resource use, reduce waste and improve process control, which strengthens the case for applying such systems in textile manufacturing\cite{rahman2024}.

In the textile industry, several existing monitoring strategies and technologies are currently deployed to track and manage energy consumption. IoT-based monitoring systems are widely used, where smart meters, current/voltage sensors and IoT gateways are installed across different machines to provide real-time data collection and cloud-based dashboards for analysis\cite{inspirisys2025}. Similarly, SCADA (Supervisory Control and Data Acquisition) and web-based systems are implemented in many textile mills to monitor parameters such as voltage, current, power and power factor, offering historical trends and mill-wide reporting features\cite{rahman2024scada}. Another approach is smart metering systems, where digital meters are placed at departmental or machine levels to give detailed breakdowns of electricity consumption, often integrated with larger energy management systems (EMS)\cite{daitan2025}. Additionally, dedicated textile-focused EMS solutions have been introduced, which combine energy monitoring with production data to generate alerts, notifications and performance insights tailored for textile operations\cite{bmsvision2025}. However, all these approaches fall under intrusive load monitoring (ILM), since they require multiple sensors, wiring and hardware installations for each machine, leading to higher costs and complex maintenance demands.

Despite their widespread use, intrusive load monitoring (ILM) systems face several limitations in textile industries. First, ILM requires multiple sensors and wiring across machines, which increases hardware costs, installation complexity and makes scaling difficult in large factories\cite{palacios2023}. Second, frequent calibration and maintenance are necessary, since sensors may degrade over time or produce inaccurate readings in harsh textile environments with dust, humidity, and vibration\cite{li2022}. Third, data collection from multiple devices creates integration challenges, as synchronizing large volumes of heterogeneous sensor data with production systems often demands advanced IT infrastructure and expertise, which many SMEs cannot afford\cite{palacios2023}.

To address the challenges associated with Intrusive Load Monitoring (ILM), Non-Intrusive Load Monitoring (NILM) has emerged as a promising alternative. NILM is a technique that estimates the energy consumption of individual appliances by analyzing the aggregate power signal from a single measurement point, such as the main supply line, without the need for dedicated sensors on each device\cite{tanoni2024}. This approach significantly reduces hardware requirements and simplifies installation processes\cite{tanoni2024}.Some of the advantages of NILM over ILM are:
\begin{itemize}
\item Cost-Effectiveness: NILM eliminates the need for multiple sensors and wiring, leading to substantial reductions in hardware and installation costs [19]\cite{onlineNILM2024}. This makes it an attractive option for industries seeking scalable energy monitoring solutions\cite{onlineNILM2024}.

\item Simplified Maintenance: With fewer physical components, NILM systems require less frequent calibration and maintenance, reducing operational downtime and associated costs\cite{nilmEnergyDisagg}.

\item Scalability and Retrofitting: NILM can be easily integrated into existing infrastructure without major modifications, allowing for scalable deployment across various industrial settings\cite{nilmEnergyDisagg}.
\end{itemize}

Non-Intrusive Load Monitoring , also referred to as energy disaggregation.Instead of directly metering each machine, NILM algorithms separate the total load signal into appliance-level components through intelligent processing of the data\cite{piccialli2019attention}. The general workflow of NILM involves several steps: data acquisition from aggregate voltage, current or power measurements; event detection to identify ON/OFF transitions or state changes; feature extraction from steady-state, transient or harmonic signatures and finally, application of machine learning or statistical models to classify appliance states and regress their power usage\cite{azizi2020event,ghosh2019improved}. Evaluation metrics such as MAE, SAE, F1-score and NDE are commonly employed to validate the effectiveness of these\begin{figure}[htbp]
\centering
\includegraphics[width=0.45\textwidth]{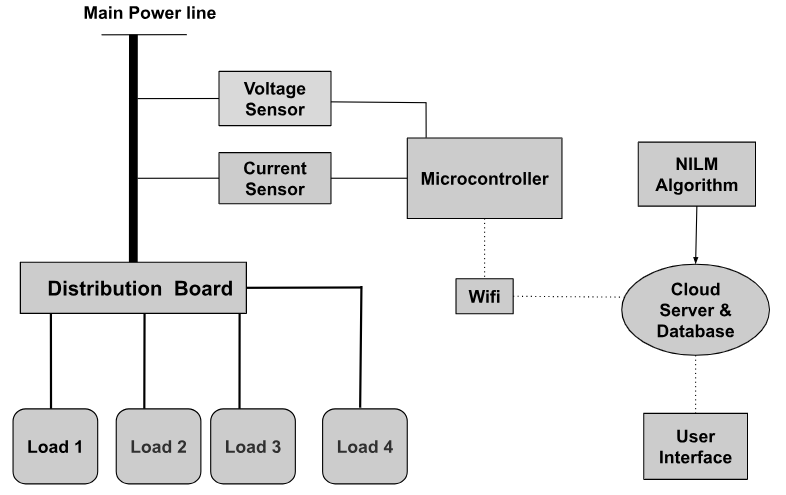}
\caption{Design Approach of the System}
\label{fig1}
\end{figure} disaggregation techniques [24]\cite{kelly2021datasets}.

For NILM systems to function effectively, certain requirements must be satisfied. First, reliable sensing equipment is needed to capture aggregate electrical signals at an appropriate sampling frequency that preserves appliance signatures—low-frequency measurements may capture steady-state features while high-frequency data allow detection of transient signatures\cite{azizi2020event}. Signal preprocessing such as noise filtering and calibration is also essential to ensure accurate analysis. Moreover, robust algorithms are required to deal with overlapping appliance usage, identical loads, and simultaneous state transitions, which are common in industrial contexts\cite{azizi2020event}. Finally, NILM depends on the availability of labeled datasets for training and validation, although recent advances leverage semi-supervised learning and data augmentation to overcome data scarcity challenges\cite{piccialli2019attention,kelly2021datasets}.

Recent advances in machine learning and deep learning have significantly improved NILM’s ability to detect appliance states and estimate their power consumption, particularly in residential environments where benchmark datasets such as REDD and UK-DALE have been widely used. However, most existing NILM research has been centered on households, with limited exploration of industrial applications such as textile manufacturing, where machines are often motor-driven, operate with high similarity and draw substantial amounts of energy. These unique characteristics pose new challenges for NILM models, particularly when attempting to distinguish between identical loads under low-frequency, real-time monitoring conditions.

This paper makes several key contributions to the field of energy monitoring and non-intrusive load monitoring (NILM), with a focus on textile industry applications:
\begin{itemize}
\item Advanced Real-Time Monitoring Framework
While existing studies have explored online or real-time NILM systems, most have been limited to residential environments and lack full integration of data collection, model deployment and monitoring. In this work, we present a complete framework that combines these components into a complete system, enhanced with remote accessibility and a user-friendly interface, making it suitable for practical deployment in industrial settings.

\item Dataset Creation - Most NILM studies rely on existing benchmark datasets such as REDD, UK-DALE and others synthetic data,  which are predominantly centered on residential appliances. In contrast, this paper introduces a new dataset created from experimentally tested loads that emulate industrial operating conditions, offering a more relevant benchmark for evaluating NILM performance in textile industry applications.

\item Evaluation of a Model on Identical Loads – The study evaluates the performance of a state-of-the-art NILM model which was trained by other types of datasets and compares with the result after deployment under the challenging scenario of identical loads. This comparison offers valuable insights into NILM effectiveness when industrial machines, such as motor-driven equipment, exhibit highly similar power signatures.

\item Revealing Challenges in NILM for Industrial Loads – The research highlights critical challenges NILM faces when applied in industrial environments, including the difficulty of distinguishing identical loads, reliance on low-frequency data and limitations posed by small-scale datasets. These findings underscore the need for more robust NILM solutions tailored to industrial applications.
\end{itemize}

\section{Related Works}
Non-Intrusive Load Monitoring (NILM) has been an active research area for decades, with works spanning from classical algorithms to modern deep learning models. Early NILM approaches relied on event-based methods, which detect appliance switching events (on/off) and extract steady-state or transient features from aggregated low-frequency signals\cite{hart1992nilm}. These techniques, though foundational, often suffered from poor scalability in multi-appliance environments. To overcome these limitations, machine learning and deep learning approaches emerged, leveraging classification and regression models to improve disaggregation accuracy. For example, convolutional neural networks (CNNs) and recurrent neural networks (RNNs) have been used to model temporal dependencies and achieve more accurate load identification compared to traditional statistical methods\cite{bonfigli2017afhmm}.

In terms of algorithms, supervised learning approaches such as SVM, Random Forest and XGBoost have been widely applied for NILM tasks\cite{kim2011unsupervised}. More recently, deep learning-based architectures including CNNs, RNNs/LSTMs and transformer-based models have achieved state-of-the-art results in energy disaggregation\cite{kelly2015neural,bonfigli2018deep}. In addition, specialized models such as locality constrained DNNs and feature-driven classifiers have been proposed to enhance the identification of similar appliances\cite{zhang2018seq2point}.

More advanced NILM models have adopted sequence-to-sequence and sequence-to-point architectures, where networks such as LSTMs learn temporal representations directly from time-series data\cite{zhang2018seq2point}. These approaches, evaluated on datasets like UK-DALE and REDD, demonstrated significant improvements in disaggregation accuracy for residential appliances\cite{zhang2018seq2point}. In particular, multi-task and hybrid frameworks were introduced to improve regression and classification simultaneously while reducing the dependence on labeled data\cite{xiong2023matnilm}.In parallel, researchers investigated sampling rates and feature representations, showing that higher-frequency data captures transient load signatures more effectively, while lower-frequency data can still achieve acceptable accuracy when combined with robust machine learning models\cite{kolter2011redd}. Datasets such as REFIT, COLD, PLAID and AMPds have also been central to benchmarking, with works testing NILM algorithms at different frequencies and feature sets (current, voltage, harmonics, power) to explore trade-offs between accuracy and computational complexity\cite{kelly2015ukdale}.

Recent studies have extended NILM toward real-time monitoring frameworks, highlighting its potential for deployment. For instance, transformer-based models such as COLD were introduced to handle concurrent loads, disaggregating up to eleven appliances operating simultaneously on high-frequency datasets\cite{kamyshev2021cold}. Similarly, hybrid classification–regression approaches have been proposed for high-frequency NILM, demonstrating real-time responsiveness with residential appliances\cite{chen2022hybrid}. A real-time NILM system leveraging uncorrelated spectral components of active power also proved that efficient performance is possible even at lower sampling rates\cite{welikala2019realtime}. These studies suggest that NILM can move from purely offline evaluation toward online and real-world applications. For instance, the study by Garcia-Marrero et al. (2025) developed a real-time NILM framework for constrained edge devices, but it was primarily tested in residential environments and did not incorporate a comprehensive user interface for remote monitoring\cite{garcia2025robust}. Similarly, Welikala et al. (2019) implemented a real-time NILM system using single active power measurements, but their approach lacked integration with real-time monitoring platforms and user interfaces\cite{welikala2019robust}.

However, NILM still faces challenges with identical or nearly identical appliances. For example, Matindife et al. investigated the classification of LED lamps with equal power ratings and reported that disaggregation accuracy deteriorates when appliances share similar steady-state features, leading to confusion in both classification and regression models\cite{matindife2020performance}. This aligns with broader findings that NILM struggles to distinguish devices with overlapping signatures, particularly under low-frequency sampling\cite{zoha2012survey}. Although industrial NILM has gained some traction, most studies remain limited compared to residential domains. For instance, Zhou et al. proposed a multiple-electrical-parameter NILM model applied to industrial equipment, where point-to-point decomposition improved disaggregation in complex environments\cite{zhou2024multiple}. 

NILM techniques have evolved from traditional event-based models to deep learning and real-time frameworks, tested across diverse datasets and sampling settings. Yet, the gap persists in handling identical industrial loads with real-time monitoring, where the performance degrades significantly. Our work seeks to bridge this gap by constructing an industrially\begin{figure}[htbp]
\centering
\includegraphics[width=0.45\textwidth]{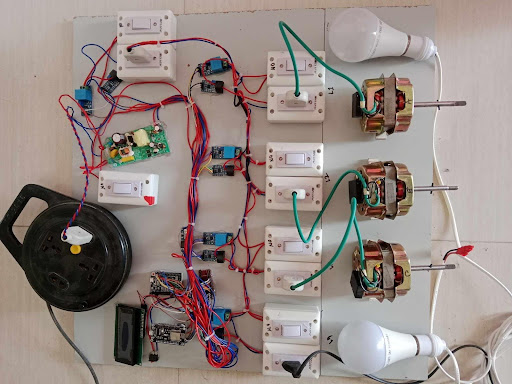}
\caption{Hardware Setup of the System}
\label{fig2}
\end{figure} motivated dataset with identical induction motors representing textile cutting machines, evaluating a state of the art model under these conditions that reveals the limitations and opportunities of NILM in such applications.

\section{Methodology}

\subsection{System Design and Framework}
The proposed system was designed to enable real-time energy monitoring and NILM-based disaggregation for textile industry loads. To emulate a small-scale cutting section, three identical 50W induction motors were connected to three different lines to represent textile cutting machines, while two additional bulb loads were included in series on another different line to introduce diversity.Replacing cutting machines with induction motors is reasonable, given that these machines are primarily motor-driven and induction motors dominate industrial usage owing to their durability, efficiency and adaptability for continuous processes making them a suitable replacement in the experimental setup\cite{singh2017textile}. Furthermore, motors account for the majority of energy consumption in textile plants, often contributing more than 70\% of total electrical energy usage\cite{bhadra2019textile,DHAYANESWARAN2014340}, making them a realistic proxy for cutting machines in NILM experimentation. The inclusion of two 15W bulb loads which are connected in series provides additional diversity in the dataset and reflects the lighting systems that typically share the same distribution line in textile facilities\cite{bhadra2019textile}. This combination of loads therefore captures both the dominant motor-driven consumption and the auxiliary lighting demand, enabling a more representative evaluation of NILM performance.

Sensing Unit: Voltage and current were measured using ZMPT101B and ZMCT103C sensors, respectively. Each load line and the main line were instrumented, ensuring complete acquisition of aggregate and individual load data for training the model.

Data Processing Unit: An Arduino Mega 2560 was used to digitize sensor signals and compute real-time electrical parameters like voltage, current and power.

Data Transmission Unit: An ESP8266 Wi-Fi module enabled wireless communication, transmitting processed data to cloud platforms for storage and analysis.

Data Storage and Disaggregation: Real-time data were stored in Google Sheets, while NILM algorithms were trained and deployed on Google Colab to perform disaggregation. The disaggregated results were fed back to Google Sheets and visualized on the user interface.

User Interfaces: A 20×4 LCD display provided local real-time readings, while the Blynk application offered remote monitoring using mobile and web.

\subsection{Dataset Creation}
The data was directly collected from the actual loads stored the readings in CSV format on a local computer. Three identical induction motors are connected in L1, L2 ,L3 and the lighting section is connected in L4.  The recorded parameters are the Voltage, Current and Power of the main line as well as the load lines. These were obtained from five pairs of sensors connected to each lines. Data collection was performed in the lab over a period of 15 days, where 16 different load state combinations were manually tested by turning the loads on and off at varying times.

In total, 180,631 samples were collected during this period. The sampling rate was not fixed; on average, a new sample was recorded approximately every 5 seconds. Although higher-frequency data generally enhances NILM performance, this rate was chosen due to practical limitations of the setup. Importantly, prior research has shown that low frequency NILM datasets ( 0.2–1 Hz) are commonly used in residential and embedded monitoring systems and many of the  research shows how do they perform on the model in terms of power disaggregation and classification.\cite{kelly2015ukdale,kolter2011redd,bonfigli2017unsupervised}.

\subsection{Model Selection}
We employed MATNILM (Multi-Appliance-Task Non-Intrusive Load Monitoring) as the core NILM algorithm to disaggregate appliance level consumption from aggregated main line power, voltage and current measurements. MATNILM leverages multi-task learning to jointly perform regression and classification across multiple devices simultaneously\cite{xiong2023matnilm}. A key innovation of MATNILM is its 2D Multi-Head Self-Attention (2DMA) mechanism, which captures temporal dependencies and inter-appliance correlations, enabling the model to recognize common co-usage patterns and improve prediction accuracy. In addition, MATNILM incorporates smart data augmentation techniques, which allow it to generalize more effectively even when the labeled training data is limited as many real-world environments have scarce labels\cite{xiong2023matnilm}.

\begin{figure}[htbp]
\centering
\includegraphics[width=0.45\textwidth]{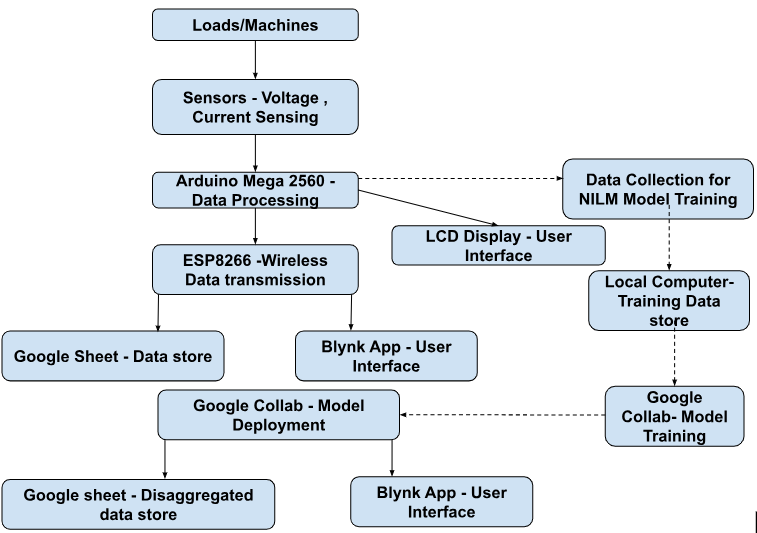}
\caption{Working Flow of System}
\label{fig4}
\end{figure}

Notably, MATNILM was originally trained and validated on the REDD  and UK-DALE dataset, both of which have relatively high sampling rates with different loads\cite{kamyshev2021cold}. This residential setting implies that the model has been proven in environments with non-identical loads and more varied appliance usage. For our project, which uses identical induction motors and operates under low or variable sampling rate , MATNILM’s capacity for learning from limited data and from datasets of lower resolution is especially desirable. 

\subsection{Dataset Preparation and Training Procedure}
The dataset collected from the hardware setup was used to train and evaluate the MATNILM model. The input features consisted of three parameters from the main line: Main Power, Main Voltage, and Main Current. The output targets were the disaggregated per-line power consumptions: L1/M1 Power, L2/M2 Power, L3/M3 Power and L4/M4 Power, corresponding to the loads connected to each line. 

For model evaluation, the dataset was divided into training, validation and testing subsets based on specific time intervals showed in \textbf{Table I}.  This method was used to keep the real picture of the data, as it reflects the actual load patterns. If the data were shuffled and split randomly, the true sequence and behavior of the loads would not be properly expressed. This strategy prioritized maximizing the size of the training set to improve generalization. 

To optimize the performance of the MATNILM model , it was trained with fine-tuned hyperparameters tailored to the collected data. The input size was fixed to 3, with a batch size of 32 and a learning rate of 0.001. The encoder–decoder hidden layer size was set to 64, and a dropout rate of 0.1 was used to mitigate overfitting. Both input and output sequence lengths were set to 864 to align with the model’s temporal window. While  Training was carried out for 5 epochs and data augmentation was applied to enhance robustness under the limited dataset scenario.
\begin{table}[htbp]
\centering
\caption{Dataset Division for Model Training and Evaluation}
\begin{tabular}{lcc}
\toprule
\textbf{Set} & \textbf{Samples (\%)} & \textbf{Dataset Duration} \\
\midrule
Training   & 146,300 (81.0\%) & [90 Hours] \\
Validation & 13,768 (7.6\%)   & [10 Hours] \\
Testing    & 20,563 (11.4\%)  & [25 Hours] \\
\bottomrule
\end{tabular}
\label{tab:dataset_split}
\end{table}
\subsection{Performance Matrices}
To evaluate the effectiveness of the proposed NILM framework, we used Mean Absolute Error (MAE), Signal Aggregate Error (SAE), F1-score and Normalized Disaggregation Error (NDE). Each metric has been widely adopted in NILM research because it captures a different aspect of disaggregation performance.

\textbf{Mean Absolute Error (MAE):} MAE measures the average magnitude of errors between predicted and actual appliance-level power. It is widely used in NILM studies because it provides a straightforward measure of instantaneous disaggregation accuracy\cite{xiong2023matnilm,kamyshev2021cold}. Mean Absolute Error is defined as: 
\begin{equation}
MAE = \frac{1}{H}\sum_{t=1}^{H} |y_t - \hat{y}_t|
\end{equation}

where $y_t$ and $\hat{y}_t$ represent actual and predicted power at time $t$, and $H$ is the total number of time steps.

\textbf{Signal Aggregate Error (SAE):} SAE evaluates the total difference between predicted and actual energy over sub-horizons, making it suitable for assessing long-term energy estimation, which is critical for billing or energy management applications \cite{xiong2023matnilm, matindife2020performance}. SAE is defined as:
\begin{equation}
SAE = \frac{1}{S}\sum_{\tau=0}^{S-1} \frac{1}{M}|y_\tau - \hat{y}_\tau|
\end{equation}

where $y_\tau = \sum_{t=1}^{M} y_{M\tau+t}$,  
$M$ is the sub-horizon length, $S$ is the number of sub-horizons, $y_\tau$ is the total actual consumption, and $\hat{y}_\tau$ is the predicted total in that sub-horizon.

\textbf{F1-Score:} Since NILM involves detecting appliance ON/OFF states, classification performance is crucial. The F1-score, the harmonic mean of precision and recall, is commonly used to handle imbalanced datasets, where appliances are OFF most of the time\cite{xiong2023matnilm,kelly2015ukdale}. It is defined as:
\begin{equation}
F1 = \frac{2 \times precision \times recall}{precision + recall}
\end{equation}
F1-score captures both false positives and false negatives, ensuring reliable evaluation of appliance state detection.

\textbf{Normalized Disaggregation Error (NDE):}NDE is a normalized measure of the squared difference between predicted and actual appliance power, allowing fair comparison across appliances with different energy scales\cite{kamyshev2021cold,hart1992nilm}. It is defined as:
\begin{equation}
NDE = \frac{\sqrt{\sum_t (y_t - \hat{y}_t)^2}}{\sqrt{\sum_t y_t^2}}
\end{equation}
NDE was chosen because it reduces bias towards high-consumption appliances, ensuring the model’s performance is fairly evaluated for both high- and low-power devices.

\begin{table}[htbp]
\centering
\caption{Performance Metrics Across Loads }
\begin{tabular}{lccccc}
\toprule
Metric & M1 & M2 & M3 & M4 & Avg \\
\midrule
MAE & 8.79 & 9.83 & 12.68 & 6.20 & 9.38 \\
SAE & 8.65 & 9.77 & 10.48 & 5.69 & 8.65 \\
F1  & 0.79 & 0.79 & 0.68 & 0.66 & 0.73 \\
NDE & 0.63 & 0.86 & 0.73 & 0.76 & 0.75 \\
\bottomrule
\end{tabular}
\label{tab1}
\end{table}
\section{Results and Insights}

\subsection{Model Evaluation}
The performance of the MATNILM model was evaluated using four standard metrics: MAE, SAE, F1-score and NDE. The results, summarized in \textbf{Table II} , show that MAE values ranged between 6.20 W (M4) and 12.68 W (M3), with an average of 9.38 W, reflecting moderate point-wise deviations between predicted and actual signals. SAE values varied from 5.69\% (M4) to 10.48\% (M3), averaging 8.65\%, suggesting that the model was able to capture overall energy consumption reasonably well, although certain loads such as M2 and M3 showed higher deviations. The F1-scores were comparatively lower, ranging from 0.66 (M4) to 0.79 (M1, M2), which indicates difficulty in detecting ON/OFF events of the machines. Finally, the NDE values were relatively high across all loads, ranging from 0.63 (M1) to 0.86 (M2), with an average of 0.75, showing that normalized errors remained significant despite capturing total energy moderately well. Overall, the \textbf{Table II} highlights that while the framework estimated aggregate energy effectively, it struggled in achieving accurate disaggregation at the per-appliance level.

To further investigate the variations observed in the evaluation scores, two sets of analyses were conducted based on machine operating conditions. The first analysis examined performance metrics with respect to the number of machines operating simultaneously. The plots in \textbf{Fig 04.} revealed that when only a single machine was active, the model performed considerably well, with low MAE and SAE values and relatively high F1-scores, indicating accurate estimation of power consumption as well as reliable detection of ON/OFF states. However, as soon as two or more machines operated concurrently, the overall performance degraded. In these cases, MAE and SAE increased due to overlapping power signatures, F1-scores dropped as the model frequently misclassified ON states, and NDE values rose sharply, highlighting the difficulty in separating aggregated signals into correct per-machine components. This trend underscores the fundamental challenge of NILM in handling identical appliances, where the aggregated profiles are nearly indistinguishable during simultaneous operation.

\begin{figure}[htbp]
\centering
\includegraphics[width=0.45\textwidth]{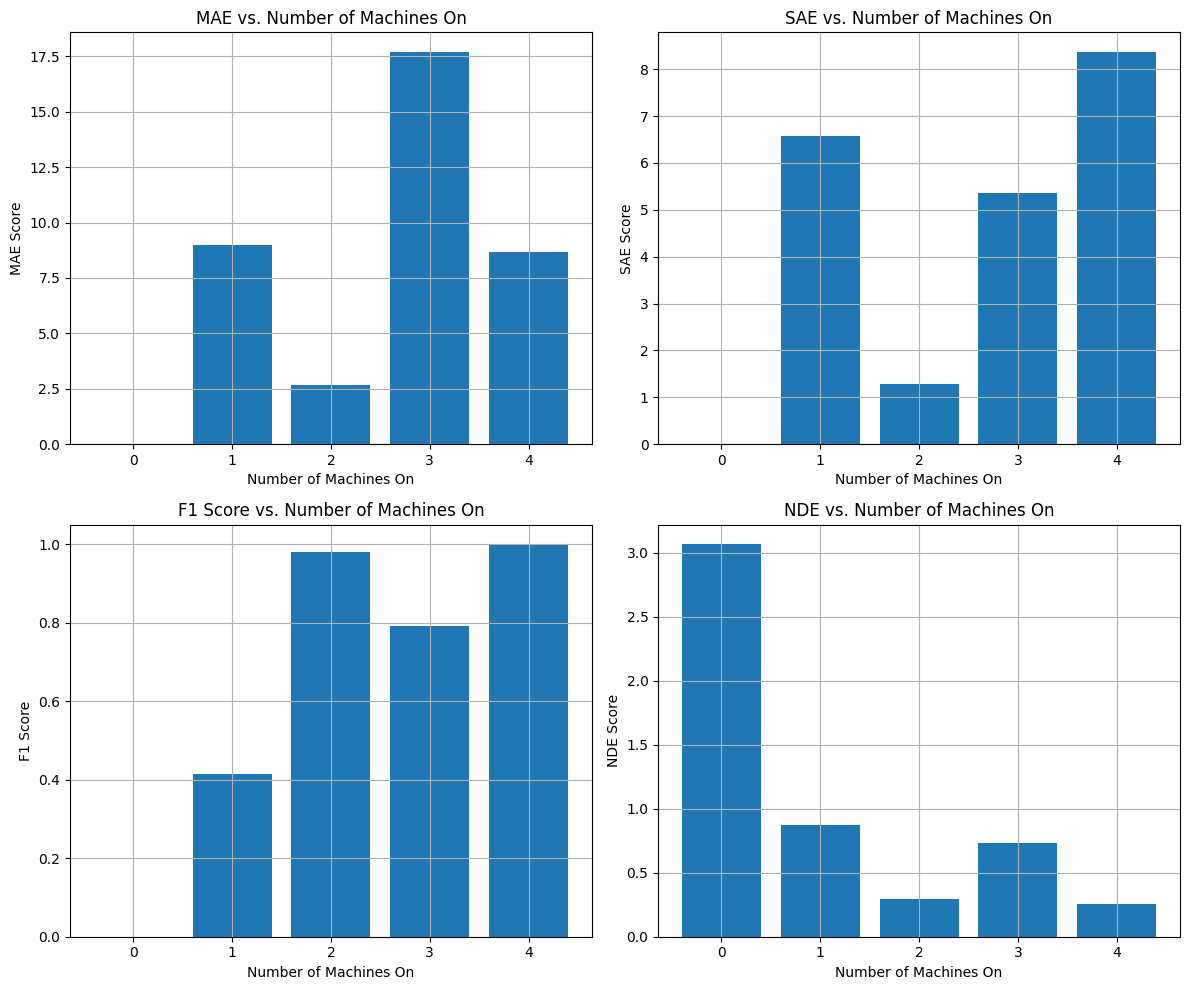}
\caption{Performance Metrics vs Number of Machines On}
\label{fig5}
\end{figure}
\begin{figure}[htbp]
\centering
\includegraphics[width=0.45\textwidth]{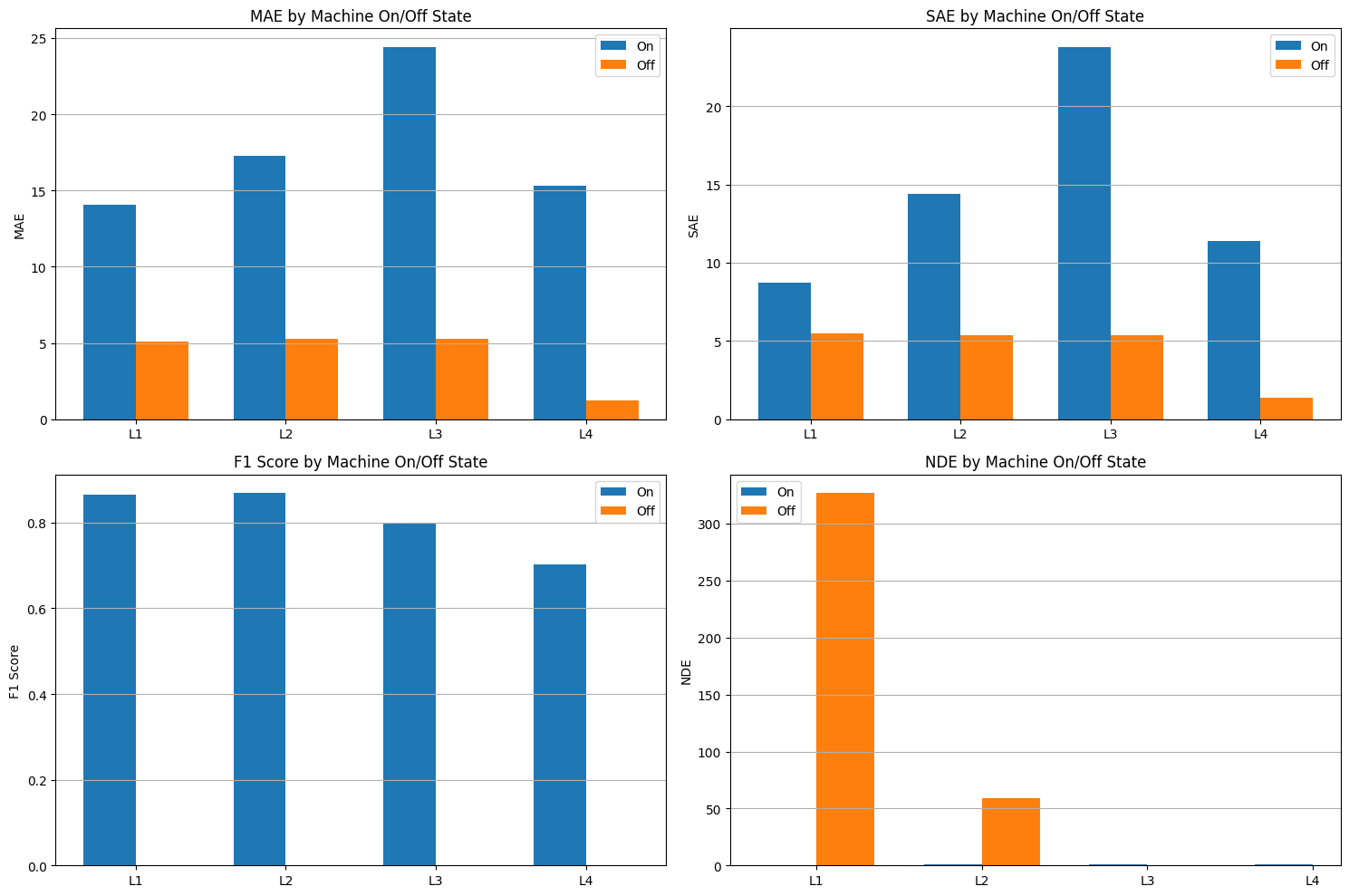}
\caption{Performance Metrics by Machine ON/OFF State}
\label{fig6}
\end{figure}
The second analysis evaluated performance based on the ON/OFF states of individual machines shown in \textbf{Fig 05.} During OFF states, the predictions were generally accurate, as inactive loads contributed no consumption, which resulted in consistently low errors across MAE and SAE. By contrast, during ON states particularly when multiple machines transitioned ON simultaneously the model often misallocated energy across appliances. This misallocation caused inflated MAE and SAE values, reduced F1-scores, and higher NDE, especially for M3 and M4. These findings are consistent with the lower F1-scores reported for these two machines in the evaluation table, confirming that ON-state overlaps among identical loads are the primary source of error. Together, these two analyses show that the model is effective in disaggregating individual loads under simple conditions but struggles significantly in complex scenarios where identical machines operate at the same time.

The actual versus predicted power consumption plots for the four monitored loads in \textbf{Fig 06.} provide visual confirmation of the performance trends reflected in the evaluation metrics. For L1, L2 and L3, which represent identical induction motor loads, the MATNILM model was poorly able to detect ON/OFF transitions and approximate steady-state trends. However, because all three machines exhibited nearly identical power signatures, the model often confused their states, leading to deviations in predicted magnitudes. This explains the moderate MAE\begin{figure}[htbp]
\centering
\includegraphics[width=0.45\textwidth]{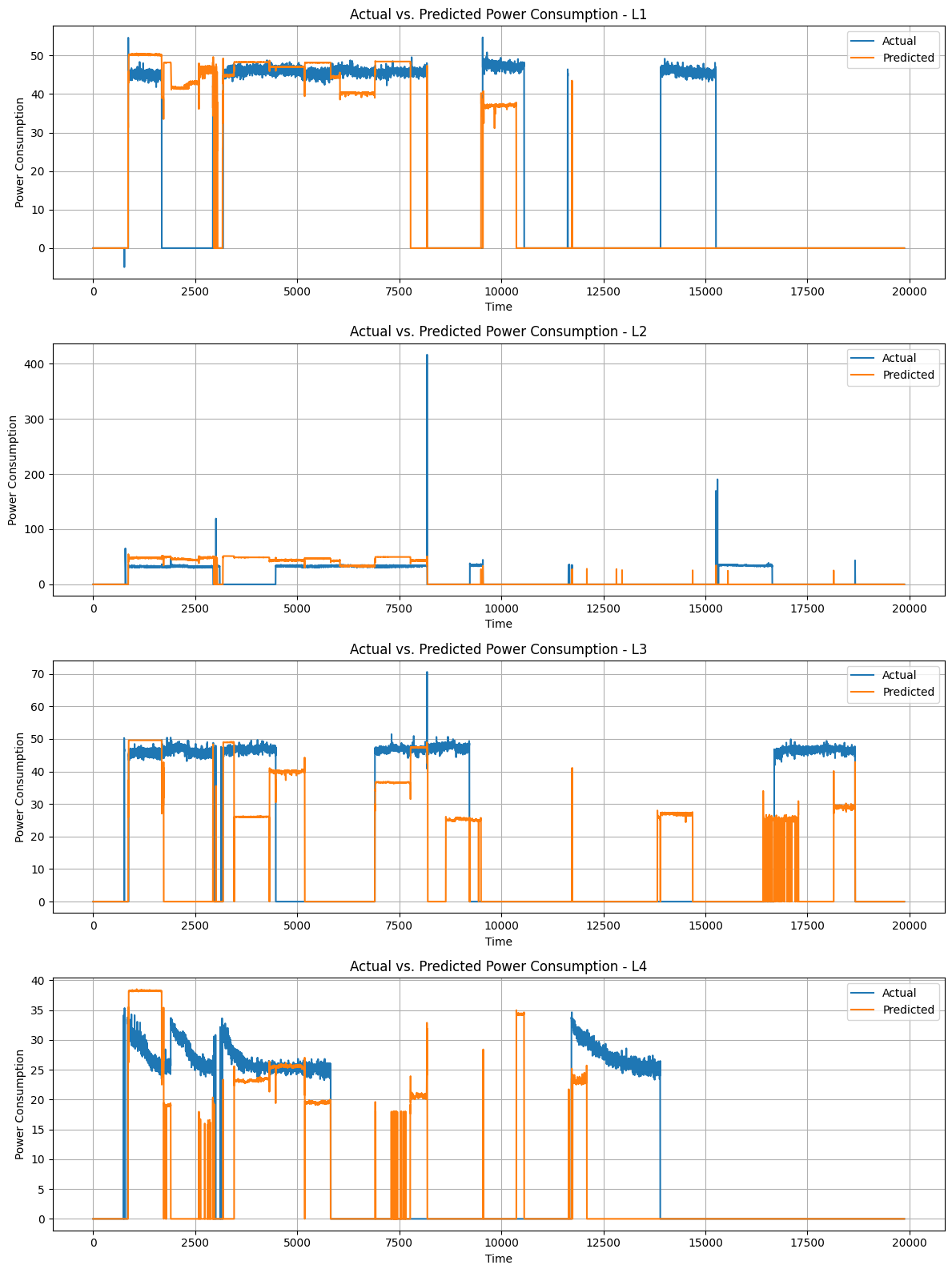}
\caption{Actual vs Predicted Power Consumption for Loads}
\label{fig7}
\end{figure}
values 9–13 W and the relatively low F1 scores 0.66–0.79, highlighting NILM’s well-known difficulty in disaggregating identical or highly similar loads.

Within the three motors, L2 showed the greatest mismatch, especially during sudden spikes and transient events, which the model consistently underestimated. These errors correspond to its higher SAE 10.5\% and worst NDE 0.86 among the loads. A likely contributing factor to this issue is the sensor readings themselves, as occasional irrelevant or noisy values may have distorted the transient signatures, leading to further inaccuracies in prediction.

For L4, non-motor load, the model tracked ON/OFF states more effectively and achieved lower MAE 6 W and SAE 5.7\%. Nonetheless, steady-state magnitudes were frequently over- or underestimated, reflected in its NDE 0.76. While easier to distinguish from the three motors, prediction accuracy was still constrained by model generalization.

The poor performance observed in both evaluation scores and actual vs. predicted plots can be attributed to three main factors. First, the monitored machines (L1–L3) were identical induction motors, resulting in highly overlapping power signatures. Since NILM algorithms rely on distinctive load features, the model struggled to differentiate the ON/OFF status and power levels of machines with nearly identical patterns. Second, the dataset was collected at a low sampling frequency, which restricted the capture of transient behaviors\begin{figure}[htbp]
\centering
\includegraphics[width=1\linewidth, height=0.55\linewidth, keepaspectratio]{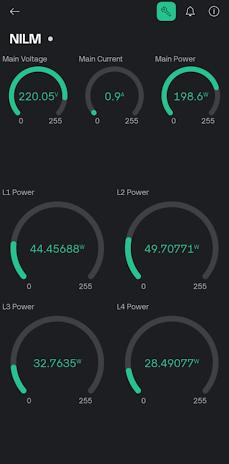}
\hfill
\includegraphics[width=1\linewidth, height=0.55\linewidth, keepaspectratio]{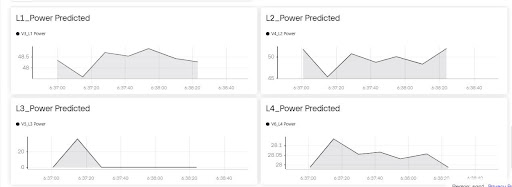}
\caption{Blynk App Interface for Real-Time Monitoring}
\label{fig:fig9}
\end{figure}
that often serve as distinguishing features between appliances. Third, the dataset size was relatively small compared to large-scale residential NILM datasets, limiting the model’s ability to generalize and accurately capture rare or abnormal operating events. Also the sensors used here gave noisy readings which affects the model ability while training and predicting. 

\subsection{System Functional Verification}
To integrate real-time monitoring with the Non-Intrusive Load Monitoring  model, the trained model was saved and deployed on Google Colab. The system is designed to fetch input data, including main voltage, main current and main power, stored in real time on a Google Sheet. Each new sample of these inputs is automatically fed into the trained model, which then disaggregates the total power into individual loads or machine level power consumption. The disaggregated values are stored in another Google Sheet and simultaneously displayed on the Blynk App, enabling remote monitoring from anywhere with internet access\cite{blynkIoT}. \textbf{Fig 07.} illustrate the remote monitoring setup. The integration with the Blynk App ensures convenient real-time access to system data from any remote location.
\begin{table}[htbp]
\centering
\caption{Performance Metrics Across Loads while Real Time Monitoring}
\begin{tabular}{lccccc}
\toprule
Metric & M1 & M2 & M3 & M4 & Avg \\
\midrule
MAE & 6.93 & 25.51 & 10.55 & 10.74 & 13.43 \\
SAE & 5.81 & 21.40 & 8.85 & 9.01 & 11.27 \\
F1  & 0.87 & 0.82 & 0.90 & 0.69 & 0.82 \\
NDE & 0.47 & 0.68 & 0.51 & 0.70 & 0.59 \\
\bottomrule
\end{tabular}
\label{tab1}
\end{table}\

\begin{figure}[htbp]
\centering
\includegraphics[width=0.45\textwidth]{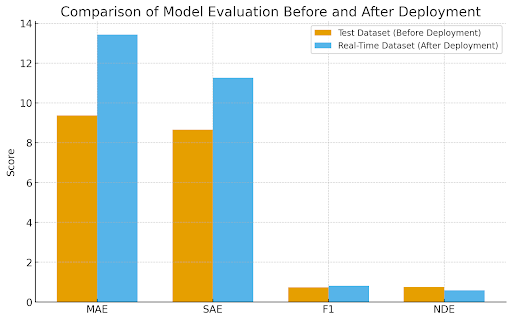}
\caption{Comparison of Disaggregation Scores Before and After Deployment}
\label{fig10}
\end{figure}

\subsection{Comparison Before and After Deployment}
After deployment, there were noticeable variations in evaluation scores with the real time dataset compared to the test dataset. In \textbf{Table III},MAE and SAE increased, likely due to differences in real-time input data quality, sensor noise, or operational variations that were not present in the controlled test dataset. F1 Score improved, showing that the model effectively detects the presence of loads, even if the predicted power values are slightly off.

NDE decreased, which can be interpreted as the model maintaining reasonable relative accuracy in power disaggregation across loads. These variations highlight that real-world deployment often introduces data shifts and noise that affect error metrics. Therefore, while the model remains functional, further tuning, retraining, or incorporating more diverse training data could improve performance consistency in real-time applications.

\section{Conclusion}
This work presented the development of a real-time NILM-based monitoring framework tailored for textile industry applications. The system successfully integrated data collection, model deployment and real-time monitoring into a unified platform, accessible remotely through the Blynk application. A new dataset was created using experimentally tested induction motors as representatives of textile cutting machines, along with additional loads for diversity. The state-of-the-art MATNILM model was evaluated on this dataset and demonstrated its ability to disaggregate appliance-level consumption under challenging industrial conditions.

Furthermore, the research revealed significant challenges NILM faces in such contexts, including the inability to reliably distinguish between identical machines, the impact of low-frequency sampling that limits transient detection and the effect of relatively small dataset size on model generalization. These limitations explain the moderate disaggregation accuracy observed in metrics such as MAE, F1 and NDE, despite acceptable aggregate performance (SAE). Sensor noise and occasional spurious values further contributed to errors, particularly in loads with transient spikes.

Looking forward, several improvements are necessary to enhance NILM performance for industrial applications. Collecting higher-frequency data would enable more effective capture of transient signatures, which are crucial for distinguishing similar machines. Expanding the dataset over longer operational periods and incorporating a wider variety of industrial loads would improve model robustness. Future work could also explore advanced deep learning architectures specifically designed for identical or highly similar loads, hybrid NILM approaches combining transient and steady-state features and integration with energy management systems to provide actionable insights for efficiency improvements


\end{document}